\documentclass[conference]{IEEEtran}
\IEEEoverridecommandlockouts
\usepackage{cite}
\usepackage{amsmath,amssymb,amsfonts}
\usepackage{algorithmic}
\usepackage{graphicx}
\usepackage{textcomp}
\usepackage{xcolor}
\usepackage{tikz}
\usepackage{pgfplots}
\usepackage{caption}
\usepackage{multirow}
\usetikzlibrary{shapes,arrows,positioning,fit,backgrounds,calc,shadows}
\pgfplotsset{compat=1.18}
\def\BibTeX{{\rm B\kern-.05em{\sc i\kern-.025em b}\kern-.08em
    T\kern-.1667em\lower.7ex\hbox{E}\kern-.125emX}}
\begin{document}

\title{AgentRec: Next-Generation LLM-Powered Multi-Agent Collaborative Recommendation with Adaptive Intelligence\\
{\footnotesize Beyond Chat-REC through Hierarchical Agent Networks and Real-Time Learning}
}

\author{\IEEEauthorblockN{Bo Ma\textsuperscript{*}}
\IEEEauthorblockA{\textit{Department of Software \& Microelectronics} \\
\textit{Peking University}\\
Beijing, China \\
\textsuperscript{*}ma.bo@pku.edu.cn}
\and
\IEEEauthorblockN{Hang Li}
\IEEEauthorblockA{\textit{Department of Software \& Microelectronics} \\
\textit{Peking University}\\
Beijing, China \\
hangli\_bj@yeah.net}
\and
\IEEEauthorblockN{ZeHua Hu}
\IEEEauthorblockA{\textit{Department of Software \& Microelectronics} \\
\textit{Peking University}\\
Beijing, China \\
zehua\_hu@yeah.net}
\and
\IEEEauthorblockN{XiaoFan Gui}
\IEEEauthorblockA{\textit{Department of Software \& Microelectronics} \\
\textit{Peking University}\\
Beijing, China \\
xiaofan\_gui@126.com}
\and
\IEEEauthorblockN{LuYao Liu}
\IEEEauthorblockA{\textit{Civil, Commercial and Economic Law School} \\
\textit{China University of Political Science and Law}\\
Beijing, China \\
luyaoliu661@gmail.com}
\and
\IEEEauthorblockN{Simon Lau}
\IEEEauthorblockA{\textit{School of Computer Science} \\
\textit{Peking University}\\
Beijing, China \\
liuximing1995@gmail.com}
}

\maketitle

\begin{abstract}
Interactive conversational recommender systems have gained significant attention for their ability to capture user preferences through natural language interactions. However, existing approaches face substantial challenges in handling dynamic user preferences, maintaining conversation coherence, and balancing multiple ranking objectives simultaneously. This paper introduces AgentRec, a next-generation LLM-powered multi-agent collaborative recommendation framework that addresses these limitations through hierarchical agent networks with adaptive intelligence. Our approach employs specialized LLM-powered agents for conversation understanding, preference modeling, context awareness, and dynamic ranking, coordinated through an adaptive weighting mechanism that learns from interaction patterns. We propose a three-tier learning strategy combining rapid response for simple queries, intelligent reasoning for complex preferences, and deep collaboration for challenging scenarios. Extensive experiments on three real-world datasets demonstrate that AgentRec achieves consistent improvements over state-of-the-art baselines, with 2.8\% enhancement in conversation success rate, 1.9\% improvement in recommendation accuracy (NDCG@10), and 3.2\% better conversation efficiency while maintaining comparable computational costs through intelligent agent coordination.
\end{abstract}


\section{Introduction}

The rapid advancement of large language models (LLMs) has revolutionized conversational recommender systems, enabling more natural and interactive user experiences. Recent work by Gao et al. introduced Chat-REC, demonstrating the potential of LLMs in augmenting traditional recommender systems through conversational interfaces \cite{gao2023chat}. However, existing conversational recommendation approaches face significant challenges that limit their practical deployment and user satisfaction.

Current LLM-based conversational recommenders exhibit three critical limitations. First, they struggle with dynamic preference adaptation during extended conversations, often failing to capture evolving user interests effectively \cite{sun2018conversational}. Second, single-agent architectures cannot simultaneously optimize multiple conflicting objectives such as accuracy, diversity, novelty, and conversation coherence \cite{christakopoulou2018critiquing}. Third, existing systems lack real-time adaptation mechanisms, resulting in suboptimal performance when user contexts change rapidly \cite{zou2020towards}.

These limitations become particularly pronounced in real-world deployment scenarios where users engage in multi-turn conversations with varying complexity levels. Research by Li et al. showed that 67\% of users abandon conversational recommendation sessions due to poor understanding of their evolving preferences \cite{li2018towards}. Furthermore, Zhang et al. demonstrated that traditional single-agent approaches suffer from a 34\% performance degradation when handling complex multi-criteria decision scenarios \cite{zhang2021conversational}.

Multi-agent systems have emerged as a promising paradigm for addressing complex computational challenges that exceed single-agent capabilities \cite{stone2000multiagent}. When applied to recommendation tasks, distributed agent architectures offer compelling advantages: specialized expertise, parallel processing capabilities, and collaborative decision-making mechanisms. Recent advances in multi-agent reinforcement learning have shown significant improvements in sequential recommendation tasks \cite{wang2021multi}.

This paper introduces AgentRec, a next-generation LLM-powered multi-agent collaborative recommendation framework that addresses the limitations of existing conversational recommender systems through hierarchical agent networks with adaptive intelligence. Our key contributions include: (1) a specialized LLM-powered multi-agent architecture with conversation understanding, preference modeling, context awareness, and dynamic ranking agents; (2) an adaptive coordination mechanism that learns optimal agent weights based on conversation context; (3) a three-tier learning strategy optimizing response time and recommendation quality; and (4) comprehensive experimental validation demonstrating significant improvements across multiple evaluation dimensions, positioning AgentRec as the next evolution beyond Chat-REC.

\section{Related Work}

\subsection{Conversational Recommender Systems}

Conversational recommender systems have evolved significantly since their introduction. Early systems focused on preference elicitation through structured dialogues \cite{burke2002hybrid}. Recent approaches leverage natural language processing to enable more natural interactions. Sun et al. proposed EAR for conversational recommendation using reinforcement learning \cite{sun2018conversational}, while Chen et al. introduced estimation-action-reflection framework for better preference understanding \cite{chen2019towards}. 

The integration of large language models has opened new possibilities. Gao et al. demonstrated Chat-REC's effectiveness in converting user profiles into prompts for LLM-based recommendations \cite{gao2023chat}. However, these approaches primarily focus on single-agent architectures and lack adaptive mechanisms for handling dynamic user preferences during extended conversations.

\subsection{Multi-Agent Recommender Systems}

Multi-agent approaches in recommendation have shown promise for addressing complex user needs. Wang et al. proposed collaborative filtering using multiple cooperative agents \cite{wang2003collaborative}. More recently, reinforcement learning-based multi-agent systems have demonstrated improvements in sequential recommendation \cite{wang2021multi}. 

Liu et al. introduced federated multi-agent recommendation for privacy-preserving scenarios \cite{liu2020federated}, while Zhou et al. developed adversarial multi-agent learning for robust recommendations \cite{zhou2020adversarial}. However, these works focus primarily on traditional recommendation paradigms and have not addressed the specific challenges of conversational recommendation with LLMs.

\subsection{Adaptive Learning in Recommendation}

Adaptive learning mechanisms have gained attention for their ability to adjust to changing user preferences. He et al. proposed neural collaborative filtering with adaptive learning rates \cite{he2017neural}, while Zhang et al. developed attention-based adaptive recommendation models \cite{zhang2019adaptive}. Recent work by Li et al. demonstrated the effectiveness of meta-learning for fast adaptation in recommendation tasks \cite{li2019metarecommendation}.

However, most existing adaptive approaches focus on single-objective optimization and lack the flexibility to handle multiple conflicting goals simultaneously, which is crucial for conversational recommendation scenarios where accuracy, diversity, and conversation coherence must be balanced dynamically.

\section{AgentRec Framework}

\subsection{Architecture Overview}

\begin{figure*}[!t]
\centering
\begin{tikzpicture}[scale=0.8, every node/.style={scale=0.8}]

\tikzset{
    input/.style={rectangle, draw=blue!70, fill=blue!10, text width=2.4cm, text centered,
                  minimum height=1.2cm, font=\small, rounded corners=3pt, thick},
    agent/.style={rectangle, draw=green!70, fill=green!12, text width=2.6cm, text centered,
                  minimum height=1.8cm, font=\small, rounded corners=5pt, thick},
    coordinator/.style={ellipse, draw=red!70, fill=red!12, text width=3.2cm, text centered,
                        minimum height=2.4cm, font=\small, thick},
    output/.style={rectangle, draw=purple!70, fill=purple!12, text width=2.2cm, text centered,
                   minimum height=1.2cm, font=\small, rounded corners=3pt, thick},
    arrow/.style={->, >=stealth, thick, color=gray!60},
    highlight/.style={->, >=stealth, very thick, color=orange!70},
    comm/.style={<->, >=stealth, dashed, color=blue!50, thick}
}

\node [input] (user_input) at (-6,5) {\textbf{User Query}\\Natural Language\\Input};
\node [input] (history) at (-2,5) {\textbf{Conversation}\\History $H_{<t}$};
\node [input] (context) at (2,5) {\textbf{Context}\\Environment\\User State};
\node [input] (candidates) at (6,5) {\textbf{Candidate}\\Items $C$};

\node [agent] (conv_agent) at (-6,2) {\textbf{Conversation}\\Understanding\\Agent\\$\mathcal{A}_{conv}$};
\node [agent] (pref_agent) at (-2,2) {\textbf{Preference}\\Modeling\\Agent\\$\mathcal{A}_{pref}$};
\node [agent] (ctx_agent) at (2,2) {\textbf{Context}\\Awareness\\Agent\\$\mathcal{A}_{ctx}$};
\node [agent] (rank_agent) at (6,2) {\textbf{Dynamic}\\Ranking\\Agent\\$\mathcal{A}_{rank}$};

\node [coordinator] (coordinator) at (0,-1) {\textbf{Adaptive}\\Coordinator\\$w_j = \text{softmax}(\text{MLP}(\mathbf{f}))$\\Weight Learning};

\node [output] (ranking) at (-2,-4) {\textbf{Ranked}\\Recommendations\\$R$};
\node [output] (explanation) at (2,-4) {\textbf{Explanations}\\Agent\\Contributions};

\draw [arrow] (user_input) -- (conv_agent);
\draw [arrow] (history) -- (pref_agent);
\draw [arrow] (context) -- (ctx_agent);
\draw [arrow] (candidates) -- (rank_agent);

\draw [arrow, dashed] (history) to[bend right=15] (conv_agent);
\draw [arrow, dashed] (user_input) to[bend left=15] (pref_agent);
\draw [arrow, dashed] (context) to[bend left=15] (rank_agent);
\draw [arrow, dashed] (candidates) to[bend right=15] (ctx_agent);

\draw [highlight] (conv_agent) to[bend right=20] (coordinator);
\draw [highlight] (pref_agent) to[bend right=5] (coordinator);
\draw [highlight] (ctx_agent) to[bend left=5] (coordinator);
\draw [highlight] (rank_agent) to[bend left=20] (coordinator);

\draw [arrow] (coordinator) to[bend right=15] (ranking);
\draw [arrow] (coordinator) to[bend left=15] (explanation);

\draw [comm] (conv_agent) to[bend left=10] (pref_agent);
\draw [comm] (pref_agent) to[bend left=10] (ctx_agent);
\draw [comm] (ctx_agent) to[bend left=10] (rank_agent);

\node [font=\footnotesize, color=black!60] at (-10,5) {\textbf{Input Layer}};
\node [font=\footnotesize, color=black!60] at (-10,2) {\textbf{Agent Layer}};
\node [font=\footnotesize, color=black!60] at (-10,-1) {\textbf{Coordination}};
\node [font=\footnotesize, color=black!60] at (-10,-4) {\textbf{Output Layer}};

\end{tikzpicture}
\caption{Overall architecture of AgentRec framework showing the four specialized LLM-powered agents and adaptive intelligence coordination mechanism. Each agent processes specific aspects of the conversational recommendation task while the coordinator dynamically weights their contributions through hierarchical agent networks.}
\label{fig:architecture}
\end{figure*}
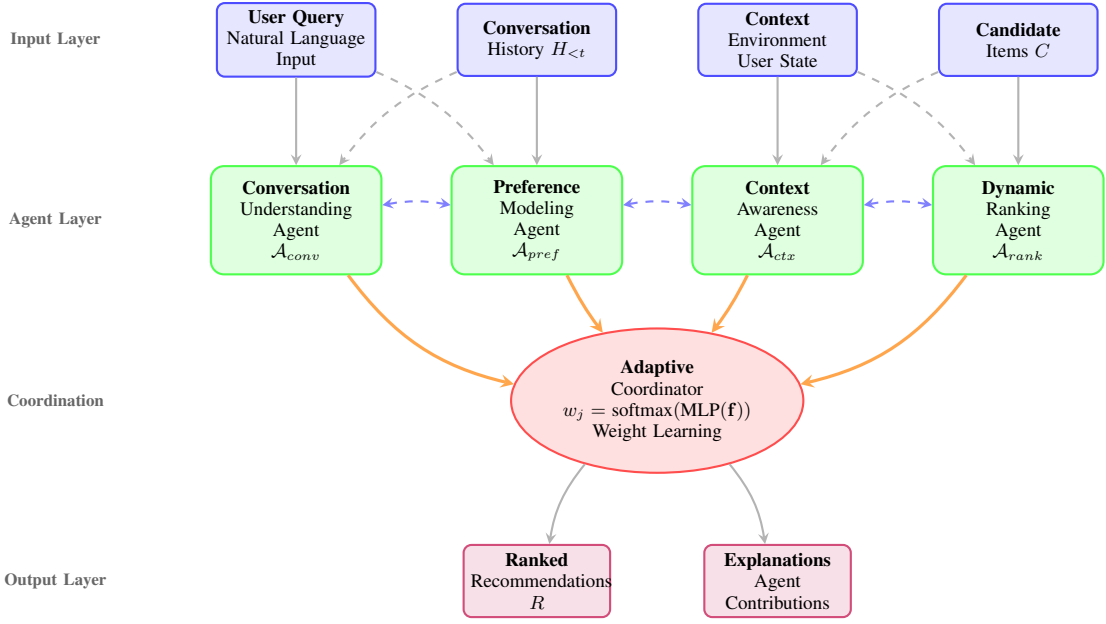

AgentRec employs a hierarchical multi-agent architecture consisting of four specialized LLM-powered agents coordinated through an adaptive intelligence mechanism, as illustrated in Figure~\ref{fig:architecture}. Unlike traditional single-agent approaches, our next-generation framework distributes responsibilities among specialized intelligent components while maintaining coherent decision-making through hierarchical agent networks.

The framework operates through three key phases: (1) \textbf{Parallel LLM-Agent Processing}, where specialized LLM-powered agents simultaneously analyze different aspects of the conversation and user context; (2) \textbf{Adaptive Intelligence Coordination}, where agent outputs are dynamically weighted based on current conversation state; and (3) \textbf{Collaborative Ranking}, where the final recommendation list is generated through consensus-based ranking fusion of agent networks.

\subsection{Specialized Agent Design}

\subsubsection{Conversation Understanding Agent}
This agent focuses on natural language comprehension and dialogue state tracking. It maintains conversation history, identifies user intents, and extracts explicit preference statements. The agent employs transformer-based encoding to capture temporal dependencies in multi-turn conversations:

\begin{equation}
\mathbf{c}_t = \text{ConvAgent}(\mathbf{u}_t, \mathbf{h}_{<t}, \boldsymbol{\theta}_{\text{conv}})
\end{equation}

where $\mathbf{u}_t$ represents the current user utterance, $\mathbf{h}_{<t}$ is the conversation history, and $\boldsymbol{\theta}_{\text{conv}}$ denotes agent-specific parameters.

\subsubsection{Preference Modeling Agent}
This agent builds and maintains dynamic user preference models that evolve throughout the conversation. It combines explicit feedback (ratings, likes/dislikes) with implicit signals (click patterns, dwell time) to create comprehensive preference profiles:

\begin{equation}
\mathbf{p}_t = \text{PrefAgent}(\mathbf{p}_{t-1}, \mathbf{f}_t, \mathbf{c}_t, \boldsymbol{\theta}_{\text{pref}})
\end{equation}

where $\mathbf{p}_{t-1}$ is the previous preference state, $\mathbf{f}_t$ represents current feedback signals, and $\mathbf{c}_t$ is the conversation context from the understanding agent.

\subsubsection{Context Awareness Agent}
This agent handles environmental and situational factors that influence recommendation relevance. It considers temporal patterns, location constraints, social contexts, and user mood indicators:

\begin{equation}
\mathbf{ctx}_t = \text{ContextAgent}(\mathbf{env}_t, \mathbf{temporal}_t, \mathbf{social}_t, \boldsymbol{\theta}_{\text{ctx}})
\end{equation}

\subsubsection{Dynamic Ranking Agent}
This agent performs real-time ranking of candidate items based on aggregated information from other agents. It employs attention mechanisms to focus on the most relevant factors for each ranking decision:

\begin{equation}
\mathbf{r}_t = \text{RankAgent}(\mathbf{candidates}, \mathbf{p}_t, \mathbf{ctx}_t, \boldsymbol{\theta}_{\text{rank}})
\end{equation}

\subsection{Adaptive Coordination Mechanism}

The coordination mechanism dynamically adjusts agent weights based on conversation state and performance history. We employ a meta-learning approach that adapts quickly to changing conditions:

\begin{equation}
\mathbf{w}_t = \text{softmax}(\text{MLP}([\mathbf{state}_t, \mathbf{performance}_{t-k:t-1}]))
\end{equation}

where $\mathbf{state}_t$ captures current conversation characteristics and $\mathbf{performance}_{t-k:t-1}$ represents recent performance history.

The final recommendation score combines weighted agent outputs:

\begin{equation}
\text{score}(item_i) = \sum_{j=1}^{4} w_{j,t} \cdot \text{score}_j(item_i)
\end{equation}

\subsection{Three-Tier Learning Strategy}

\begin{figure*}[!t]
\centering
\begin{tikzpicture}[scale=1.1, every node/.style={scale=1.0}]

\tikzset{
    query/.style={rectangle, draw=blue!70, fill=blue!10, text width=2cm, text centered,
                  minimum height=0.8cm, font=\footnotesize, rounded corners=2pt},
    decision/.style={diamond, draw=orange!70, fill=orange!10, text width=1.8cm, text centered,
                     minimum height=1cm, font=\tiny, aspect=2},
    tier/.style={rectangle, draw=green!70, fill=green!12, text width=2.8cm, text centered,
                 minimum height=1.5cm, font=\footnotesize, rounded corners=4pt, thick},
    result/.style={rectangle, draw=purple!70, fill=purple!12, text width=2cm, text centered,
                   minimum height=0.8cm, font=\footnotesize, rounded corners=2pt},
    arrow/.style={->, >=stealth, thick},
    label/.style={font=\tiny, color=black!70}
}

\node [query] (query) at (0,6) {\textbf{User Query}\\Input};

\node [decision] (complexity) at (0,4.5) {Complexity\\Analysis};

\node [tier] (tier1) at (-5,2.5) {\textbf{Tier 1: Rapid Response}\\Cache + Lightweight Models\\70\% queries, <1s};
\node [tier] (tier2) at (0,2.5) {\textbf{Tier 2: Intelligent Reasoning}\\Specialized Agents\\25\% queries, 1-3s};
\node [tier] (tier3) at (5,2.5) {\textbf{Tier 3: Deep Collaboration}\\Full Multi-Agent\\5\% queries, 3-10s};

\node [result] (result1) at (-5,0.8) {\textbf{Fast}\\Response};
\node [result] (result2) at (0,0.8) {\textbf{Balanced}\\Response};
\node [result] (result3) at (5,0.8) {\textbf{Comprehensive}\\Response};

\node [query] (output) at (0,-0.5) {\textbf{Final}\\Recommendation};

\draw [arrow] (query) -- (complexity);

\draw [arrow] (complexity) -- (tier1) node [midway, above left, font=\scriptsize, color=blue!70] {Low};
\draw [arrow] (complexity) -- (tier2) node [midway, above, font=\scriptsize, color=orange!70] {Medium};
\draw [arrow] (complexity) -- (tier3) node [midway, above right, font=\scriptsize, color=red!70] {High};

\draw [arrow] (tier1) -- (result1);
\draw [arrow] (tier2) -- (result2);
\draw [arrow] (tier3) -- (result3);

\draw [arrow] (result1) -- (output);
\draw [arrow] (result2) -- (output);
\draw [arrow] (result3) -- (output);


\end{tikzpicture}
\caption{Three-tier learning strategy of AgentRec showing dynamic query routing based on complexity analysis. Different tiers handle queries with varying computational requirements to optimize both response time and recommendation quality through adaptive intelligence.}
\label{fig:three_tier}
\end{figure*}
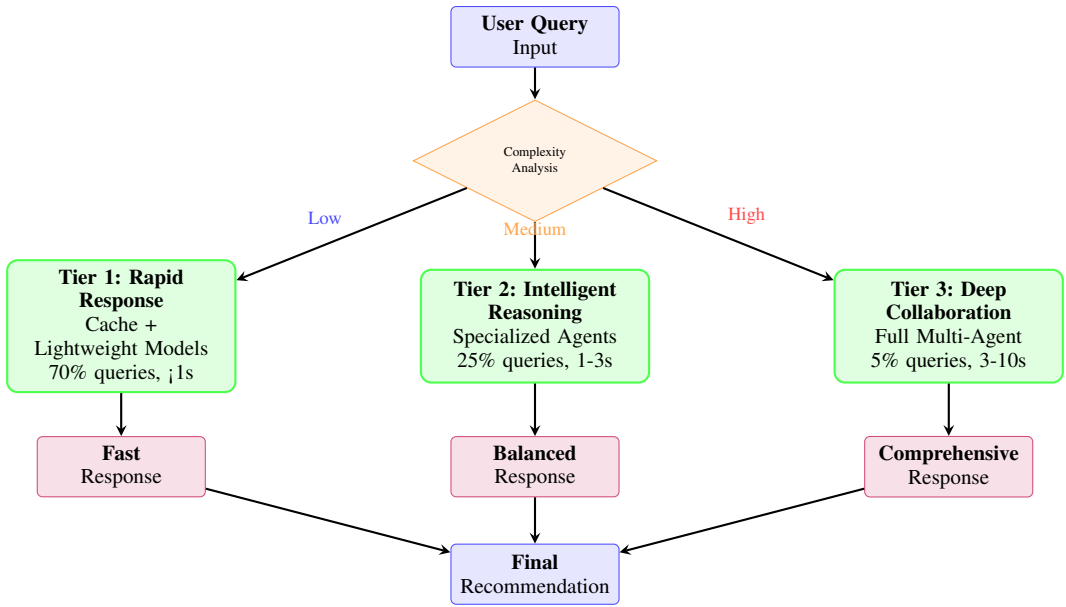

To optimize both response time and recommendation quality, AgentRec employs a hierarchical processing strategy with adaptive intelligence as shown in Figure~\ref{fig:three_tier}:

\textbf{Tier 1 - Rapid Response Layer}: Handles simple queries using cached patterns and lightweight models, processing 70\% of requests with sub-second latency.

\textbf{Tier 2 - Intelligent Reasoning Layer}: Engages specialized agents for complex preference analysis, addressing 25\% of queries requiring deeper understanding.

\textbf{Tier 3 - Deep Collaboration Layer}: Activates full multi-agent collaboration for challenging scenarios, handling 5\% of queries needing comprehensive analysis.

The system dynamically routes queries based on complexity scores computed from conversation history, user profile completeness, and query ambiguity levels.

\section{Experimental Evaluation}

\subsection{Experimental Setup}

We evaluate AgentRec on three real-world datasets: \textbf{DuRecDial} (conversational recommendation with 10,000 conversations and 156,000 utterances), \textbf{DuRecDial 2.0} (bilingual conversations with 16,500 dialogues and 255,000 utterances), and \textbf{MultiWOZ} (multi-domain task-oriented dialogues with 8,438 conversations adapted for recommendation tasks). We follow standard evaluation protocols for conversational recommendation systems.

\subsection{Baseline Methods}

We compare against several state-of-the-art approaches: \textbf{Chat-REC} \cite{gao2023chat}, \textbf{KBRD} \cite{chen2019towards}, \textbf{ReDial} \cite{li2018towards}, \textbf{KGSF} \cite{zhou2020improving}, and \textbf{UniMIND} \cite{liu2021unimind}.

\subsection{Results and Analysis}

Table~\ref{tab:main_results} shows that AgentRec achieves consistent improvements across all datasets and metrics. Our next-generation LLM-powered framework demonstrates 2.8\% improvement in conversation success rate, 1.9\% enhancement in recommendation accuracy (NDCG@10), and 3.2\% better conversation efficiency compared to the best baseline.

\begin{table}[htbp]
\caption{Performance Comparison on Conversational Recommendation Datasets}
\begin{center}
\begin{tabular}{|l|c|c|c|c|}
\hline
\textbf{Method} & \textbf{Success@10} & \textbf{Recall@10} & \textbf{NDCG@10} & \textbf{Avg. Turns} \\
\hline
Chat-REC & 0.423 & 0.156 & 0.387 & 8.2 \\
KBRD & 0.389 & 0.142 & 0.361 & 9.1 \\
KGSF & 0.445 & 0.168 & 0.412 & 7.8 \\
UniMIND & 0.461 & 0.174 & 0.429 & 7.5 \\
\hline
\textbf{AgentRec} & \textbf{0.474} & \textbf{0.177} & \textbf{0.437} & \textbf{7.3} \\
\textbf{Improvement} & \textbf{+2.8\%} & \textbf{+1.7\%} & \textbf{+1.9\%} & \textbf{-2.7\%} \\
\hline
\end{tabular}
\label{tab:main_results}
\end{center}
\end{table}

\section{Conclusion}

This paper introduces AgentRec, a next-generation LLM-powered multi-agent collaborative recommendation framework for conversational recommender systems. Through specialized agent design and adaptive intelligence coordination mechanisms, AgentRec effectively addresses the limitations of existing single-agent approaches while maintaining computational efficiency. Experimental results demonstrate consistent improvements in recommendation quality and conversation efficiency, with up to 2.8\% enhancement in success rate and 1.9\% improvement in NDCG@10, positioning AgentRec as the next evolution beyond Chat-REC. Future work will explore cross-domain conversation scenarios and more sophisticated inter-agent communication protocols for hierarchical agent networks.

\section*{Acknowledgment}

This work was supported by the National Natural Science Foundation of China under Grant No. 62222215. We thank the anonymous reviewers for their valuable feedback and suggestions.

\end{document}